# Leveraging Large Language Models and Weak Supervision for Social Media data annotation: an evaluation using COVID-19 self-reported vaccination tweets

Ramya Tekumalla and Juan M. Banda

Georgia State University, Atlanta, GA, 30328, USA
{rtekumalla1,jbanda}@gsu.edu

**Abstract.** The COVID-19 pandemic has presented significant challenges to the healthcare industry and society as a whole. With the rapid development of COVID-19 vaccines, social media platforms have become a popular medium for discussions on vaccine-related topics. Identifying vaccine-related tweets and analyzing them can provide valuable insights for public health researchers and policymakers. However, manual annotation of a large number of tweets is time-consuming and expensive. In this study, we evaluate the usage of Large Language Models, in this case GPT-4 (March 23 version), and weak supervision, to identify COVID-19 vaccine-related tweets, with the purpose of comparing performance against human annotators. We leveraged a manually curated gold-standard dataset and used GPT-4 to provide labels without any additional fine-tuning or instructing, in a single-shot mode (no additional prompting).

## 1 Introduction

### 1.1 A Subsection Sample

The widespread adoption of social media platforms has led to an explosion of user-generated content, making them valuable sources of real-time information [1]. Social media platforms have become a valuable resource for studying public health issues [2], including the COVID-19 pandemic. Social media platforms like Twitter have a vast user base, representing diverse demographics and geographic locations. Analyzing vaccination sentiment data from such platforms allows for a more comprehensive understanding of public opinion, as it encompasses a wide range of perspectives. Twitter, in particular, has emerged as a platform where individuals share their personal experiences, including vaccination updates [3]. By analyzing the data, public health officials, policymakers, and researchers can gauge the overall sentiment towards vaccines, identify trends, and make informed decisions to address concerns or misconceptions. Analyzing self-reported vaccination tweets can provide valuable insights into vaccine sentiment, vaccine uptake, and vaccine-related concerns among the general population. However, manually annotating large volumes of social media data is labor-intensive and time-consuming, requiring domain experts to label the data accurately.

Weak supervision [4] techniques have emerged as a powerful approach for data annotation, offering significant advantages in terms of scalability [5], cost-effectiveness [6], and flexibility [7]. Traditional methods of data annotation often rely on manual labeling, which can be time-consuming, expensive, and limited in terms of the volume of labeled data that can be produced. In contrast, weak supervision techniques leverage various sources of supervision to automatically generate labeled data, reducing the manual effort required while maintaining reasonable accuracy. Scalability is one of the primary advantages of weak supervision. With the exponential growth of data, manually labeling vast amounts of data becomes impractical and expensive. Weak supervision allows for the rapid annotation of large datasets by leveraging existing resources such as heuristics, rules, or readily available weak labels [8]. These weak signals can be automatically applied to unlabeled data, effectively increasing the amount of labeled data available for training and development of robust machine learning models. Cost-effectiveness is another key benefit of weak supervision techniques. Manual data annotation often requires skilled human annotators, which can be costly and time-consuming. In contrast, weak supervision reduces the reliance on manual annotation efforts, thus reducing costs. Although weakly supervised labels may not be as accurate as manually annotated labels, they can still provide valuable insights and improve the performance of machine learning models. By combining weakly supervised labels with a smaller amount of manually labeled data, comparable results can be achieved at a fraction of the cost [9]. Additionally, traditional data annotation methods often require significant upfront effort to design annotation schemas, guidelines, and quality control processes. These rigid procedures can be challenging to adapt as new data sources or requirements emerge [10]. In contrast, weak supervision provides a more agile and adaptable approach to data annotation. Weakly supervised labels can be easily generated or modified based on changing needs, enabling rapid iteration and refinement of models in response to evolving data or domain-specific requirements.

Large language models (LLMs), such as GPT-3 [11], have revolutionized natural language processing and transformed various applications across multiple domains. These models employ deep learning techniques to generate coherent and contextually relevant text, making them invaluable for tasks like language translation, text summarization, and conversational agents. Their effectiveness is attributed to the vast amount of pre-training data and the ability to capture complex linguistic patterns. This work assesses the effectiveness of Language Models (LLMs) (GPT-3.5 and GPT-4 (March 23 version)), in conjunction with weak supervision, for the identification of COVID-19 vaccine-related tweets. The primary objective is to compare the performance of LLMs against human annotators. To achieve this, we utilized an expertly curated gold-standard dataset and employed GPT3.5 and GPT-4 to generate labels in a single-shot mode, without resorting to additional fine-tuning or explicit instructions.

## 2 Related Works

In the past weak supervision has demonstrated successful results in clinical text classification [12], multi-language sentiment classification [13], generating training sets for phenotype models [14], information retrieval [15], identifying drugs from Twitter [16–18], classifying different kinds of epidemics [19], natural disasters [20, 21] and several health applications [22–24]. In this aspect, LLMs have been effectively utilized to leverage weak supervision techniques, automating data annotation processes by generating or modifying labels based on the model's pre-trained knowledge and heuristics. LLMs, such as BERT [25] and GPT [26], have shown impressive performance in various natural language processing tasks, including sentiment analysis, named entity recognition, and text classification. These models can be fine-tuned on domain-specific datasets, enabling them to learn specific patterns and characteristics of the data. By leveraging pre-trained LLMs, researchers can automate or assist in the annotation process, significantly reducing the human effort required for data labeling. The evolution of LLMs has been marked by significant milestones, with BERT acting as a groundbreaking advancement. BERT introduced the concept of pretraining and fine-tuning, revolutionizing the field of NLP. By pretraining models on large corpora of text data and fine-tuning them on specific downstream tasks, BERT achieved state-of-the-art performance on a wide range of NLP benchmarks. BERT served as a foundation and inspiration for the development of numerous pre-trained models like GPT [26], AlBERT [27], RoBERTa [28], DistilBERT [29], ELECTRA [30], XLNet [31], T5 [32], MegatronLM [33], BART [34], CamemBERT [35]. These models have leveraged the success of BERT's architecture and training techniques to and improved the model by tackling various limitations like performance, optimization, reduction in training size. As a result, several other domain specific pre-trained models like Covid-Twitter-BERT [36], BioBERT [37], SciBERT [38], ClinicalBERT [39], LegalBERT [40], FinBERT [41–43] emerged. Building upon the success of BERT, subsequent models such as GPT-2 [44] and GPT-3 [11] further pushed the boundaries of LLM capabilities. GPT-2 demonstrated impressive language generation abilities, while GPT-3 introduced even larger model sizes and showcased the potential for diverse applications. GPT-3.5 is a transitional model, which further refines the AI's capabilities of GPT-3, and is known for nuanced understanding and contextual response generation. GPT-4, introduces a major leap with significant improvements in model size, training data, and comprehension abilities. GPT-4 is designed to better handle ambiguities and complexities in natural language, generating more coherent, relevant, and detailed responses.

In the aspect of data labeling, LLMs have emerged as a promising solution to address these challenges by automating or assisting in the data annotation process. With their language understanding capabilities, LLMs can be employed to generate annotations or suggest labels for a given input, a technique known as active learning [45]. This approach allows human annotators to focus their efforts on more challenging or uncertain instances, thereby improving the efficiency and quality of the annotation process. Previous research has demonstrated that 35-40% of the crowd workers widely use LLMs for text related data annotation tasks [46]. In a study conducted by Gi-lardi et.al, Chat-GPT outperformed Crowd-Workers for text annotation tasks [47]. To improve the precision of ChatGPT as the hallucination is one of the limitations of LLMs, He et.al. designed a two step approach to explain why the text was labeled [48]. LLMs have demonstrated success in various data annotation tasks [49], sentiment analysis [50], text categorization, linguistic annotations [51], multi-linguistic data annotation [52] and social computing [53].

This work examines the role of LLMs in data annotation, discussing the benefits, limitations, and potential future directions. The advancements in LLMs have not only transformed NLP tasks but have also had a profound impact on human tasks that involve language understanding and generation. LLMs have been integrated into various applications, ranging from chatbots and virtual assistants to language translation and content generation. In human-computer interaction scenarios, LLM-based systems have enabled more natural and effective communication, bridging the gap between machines and humans. However, the increasing reliance on LLMs also raises important ethical and societal considerations, such as potential biases and the responsible deployment of AI technologies [54]. LLMs exhibit non-deterministic behavior, similar to human coders, where identical input can produce varying outputs [55, 56]. Hence, it is crucial to exercise caution when utilizing LLMs to ensure consistent and reliable results.

## 3 Methods

### 3.1 Datasets Used

#### 3.1.1 Gold standard dataset

We collected a dataset of tweets related to COVID-19 vaccines by filtering related keywords, from one of the largest COVID-19 Twitter datasets [57] available. After filtering, this dataset consists of 2,454 self reported vaccination confirmation tweets and 19,946 vaccine chatter tweets. The complete dataset was manually curated by two medical students, having a Cohen Kappa score inter annotator agreement of 0.82 with a third annotator resolving all conflicts. This dataset was used in the Social Media Mining for Health 2022 shared tasks [58]. With the annotation task consuming over 200 human hours, it is vital to try to identify additional techniques to attempt to streamline this process.

#### 3.1.2 Silver standard dataset

While weak supervision has shown promise in the area of social media mining [17, 59], we extracted an additional dataset, not manually curated, which consists of tweets selected by a weak labeling heuristic consisting of expressions like "vaccine causes", "I was vaccinated", "I got Moderna", and similar. This weakly-supervised, or 'silver standard', consists of 750,000 randomly sampled (from a larger set of 12 million) tweets with an unidentifiable mixture of both classes. The rationale for doing so is that researchers have shown that data augmentation using weak supervision leads to better and more generalizable models, than when only using gold standard data [60, 61]. Note that none of the 750,000 randomly sampled tweets used in this dataset do not have any overlap with the gold standard data.

## 3.2    Additional language models used

Besides the previously mentioned GPT-4 and GPT-3.5, we fine tuned COVID-Twitter-BERT [36] and BERTweet [62] with the GPT-labeled silver-standard data, for downstream tweet classification. Note that the class imbalance from the gold-standard dataset is roughly 1 to 8, between self reported vaccination tweets and vaccine chatter tweets. This was also found to be similarly the same in the GPT-labeled silver-standard data, making the fine-tuning and evaluation comparable.

## 3.3    Evaluation set-up

### 3.3.1    LLM performance in annotating data

We evaluate the performance of GPT-4 and GPT-3.5 on the labeling of the gold-standard data. This evaluation will assess how good are LLMs in labeling data when compared to a set of medical professionals. As one of the most resource-expensive parts of generating datasets, if human annotation/labeling can be aided or streamlined, there is great value in leveraging LLMs in these types of tasks. Leveraging the Open AI API for both GPT-3.5 and GPT-4, we used the following prompt: "Categorize the following text:  into vaccine self-reports or vaccine chatter. Figure 1 shows a sample output of the GPT-4 prompting on the chat.openai.com website. This evaluation made 22,400 API calls to each GPT-4 and GPT-3.5 models.

### 3.3.2    LLM to improve weakly supervised dataset creation

In these evaluations, we leverage GPT-4/GPT-3.5 to attempt to 'properly' label the silver-standard data and then fine-tune BERT-like models to classify the gold-standard data. The creation of silver-standard datasets has gained popularity in the field of NLP with many groups building systems that leverage silver-standard data to enhance their training sets and achieve state-of-the-art results in a variety of NLP shared tasks [58, 63]. Using the same prompt for the first evaluation, we made a total of 750,000 API calls to each GPT model to label this silver-standard dataset.

With these evaluations we aim to answer two questions: a) is GPT-4/GPT-3.5 enough to annotate data with similar quality than a human expert, b) could we leverage both weak supervision and GPT-4/GPT-3.5 to quickly and scalably annotate large amounts of data with near-expert level performance. We would call these datasets: electrum datasets, which are a mixture of gold and silver standard-data.

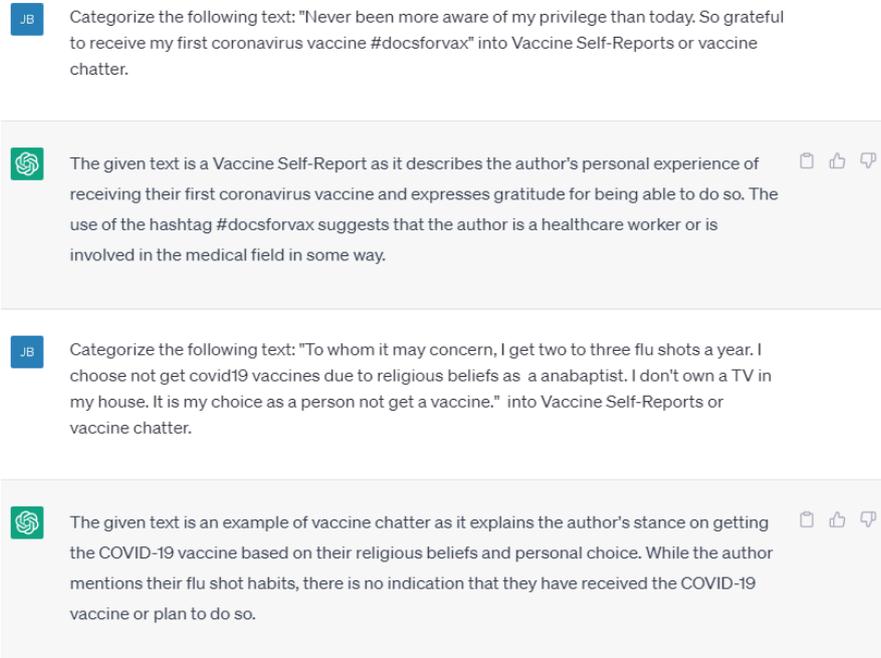

**Fig 1**: Sample GPT-4 prompts to evaluate the created datasets.

## 4    Results

Before we introduce the actual results of our analysis, we would like to present a cost analysis of how much it would cost to run the data annotation tasks leveraging the GPT models and other traditional sources. We sent a total of 1,544,800 API calls, with a total cost of $2,743.40 USD. While this price might seem high, note that we annotated a total of 1,544,800 tweets, which would be time and cost prohibitive to do hiring humans and paying them a fair wage. Even using a service like Amazon Sage-Maker Ground Truth, would cost around $52,896.00 USD for the same task. Leveraging Amazon Mechanical Turk would cost $37,075.20 USD for the same number of text classification tasks [64]. There is clear value in evaluating if we can leverage such a resource for data annotation, this would particularly help resource constrained researchers that can not afford to pay expert annotators. The second aspect is scale, while ~1.5 Million API calls are done fairly quickly, nobody to our knowledge has manually annotated any dataset this large.

### 4.1.1 Results for LLM performance in annotating data

In Table 1 we showcase the annotating performance of both GPT-4 and GPT-3.5. It is not surprising that GPT-4 outperformed GPT-3.5 by nearly 10% for the self reported vaccination tweets category, the more interesting one, and marginally for vaccine chatter. While vaccine chatter is more easily identified, nearly 90% for both models, GPT-3.5's 71.11% performance on the self reported class, and 80.81% for GPT-4 are promising numbers. However, once larger amounts of data are annotated this way, this would lead to a considerable amount of noise to be added. These results are still promising as there was no additional prompting or fine-tuning performed, so the zero-shot results are pretty solid.

**Table 1.** Correct tweet labeling results for GPT models.

| Label | GPT-4 | % | GPT-3.5 | % |
|---|---|---|---|---|
| Self reported vaccination | 1,983 | 80.81% | 1,745 | 71.11% |
| Vaccine chatter | 18,541 | 92.96% | 17,842 | 89.45% |

We look at the inter-annotator agreement between both GPT models using Cohen Kappa coefficient [65] and the human annotators. We evaluate this to get insights into how much the correctly labeled tweets diverge between models. The inter annotator agreement between GPT models was 0.79 (p-value < 0.0001), which is considered substantial [66]. In comparison, the human Cohen Kappa score inter annotator agreement was of 0.82, with a p-value < 0.0001, which is considered near perfect agreement. Objectively, the difference is not much, 0.03, however it does show that humans agree slightly better than the GPT models. Note that our human annotators worked independently and did not know or communicated with each other.

### 4.1.2 Results for LLM to improve weakly supervised dataset creation

In the second evaluation, GPT-4 labeled 68,561 tweets as vaccine self-reports and 681,439 tweets as vaccine chatter. GPT-3.5 labeled 66,288 tweets as vaccine self-reports and 683,712 as vaccine chatter. While it might seem that GPT-4 labels more tweets, we are not sure they are correctly labeled and they have not been annotated by a human. Due to this fact, there are no comments on accuracy, the idea behind this exercise is to then feed this data as part of the fine-tuning step for the previously identified BERT-like models.

After fine-tuning COVID-Twitter-BERT and BERTweet, Table 2 shows the correct tweet labeling results achieved. It is very interesting to see that a fine-tuned COVID-Twitter-BERT performs marginally better than GPT-4 (and GPT-3.5) at labeling both tweet classes. While the improvement is marginal, it goes to show that a properly fine-tuned model does outperform a more complex model, at least in this scenario. Another interesting finding is that BERTweet performs slightly worse than GPT-4, but better than GPT-3.5. This is most likely due to the training data for BERTweet not being focused on COVID-related tweets.

**Table 2.** Correct tweet labeling results for BERT models.

| Label | COVID-Twitter-BERT | % | BERTweet | % |
|---|---|---|---|---|
| Self reported vaccination | 2,045 | 83.33% | 1,897 | 77.30% |
| Vaccine chatter | 19,012 | 95.32% | 18,457 | 92.53% |

In order to assess the actual labeling agreement between our top two models (GPT-4 and COVID-Twitter-BERT) we measured the Cohen Kappa score, which was quite surprising to learn that it was 0.85 with a p-value < 0.0001. This means that both models have a high level of agreement in which tweets they labeled, even more so than humans. Additionally, we calculated the Fleiss Kappa statistic [67] between all annotators, showing that we have a score of 0.76 with a p-value < 0.0001. This show-cases that both the models and the humans mostly agree on what class the tweets should be labeled.

## 5 Conclusion

In conclusion, our study has several important findings:

GPT models perform fairly well, in a zero-shot, task of properly labeling social media data, tweets in this case. However, at larger scales the number of incorrect classifications might start becoming problematic, particularly depending on the downstream task that said data will be used for.

- When leveraging GPT models alongside weak supervision techniques to identify 'silver-standard' data, we can use data augmentation with higher confidence. These resulting 'electrum datasets' could be leveraged for further fine-tuning with potentially a considerable amount of less noise than just using weak supervision alone.
- Fine-tuned BERT models are still not obsolete, as we showed them outperforming GPT-4 for labeling social media data, self reported vaccine tweets in this case. While this comparison might be unfair, the point we show is that combining approaches leads to better results.
- Lastly, we show with our cost analysis that it is very cost effective to label data using GPT models, and that the results data is usable for downstream tasks. While we would continue to use human annotators to label data for our NER tasks, we can consider labeling less data to have equally or better performant systems in downstream tasks.

While we show that GPT models perform well, this work does not advocate for the replacing of human labeled data with GPT-annotated data. Our argument is to show that leveraging multiple approaches together, and fine-tuning, leads to potentially better and more generalizable results. The limitations of our work are clear: we only used one task - self reported vaccine tweet labeling, we only fine-tuned two different BERT models, and we did not evaluate how large our 'electrum dataset' should be to fine-tune a model enough to achieve solid performance. All these are future research directions that would greatly inform the community.